%% file: edcc20_camera_ready.tex
\documentclass[conference]{IEEEtran}
\ifCLASSINFOpdf
\else
\fi
\usepackage{cite}
\usepackage{amsmath,amssymb,amsfonts}
\usepackage{algorithmic}
\usepackage{graphicx}
\usepackage{textcomp}
\usepackage{xcolor}
\def\BibTeX{{\rm B\kern-.05em{\sc i\kern-.025em b}\kern-.08em
    T\kern-.1667em\lower.7ex\hbox{E}\kern-.125emX}}
    
\usepackage{cite}
\usepackage{subfig}
\usepackage{pifont}
\let\oldding\ding
\renewcommand{\ding}[2][1]{\scalebox{#1}{\oldding{#2}}}
\newtheorem{Tlemma}{Lemma}

\newtheorem{Tdef}{Definition}

\usepackage{xcolor}
\usepackage[linesnumbered,ruled]{algorithm2e}
\usepackage{threeparttable}
\usepackage{multirow}

\SetCommentSty{mycommfont}    
    
\newcommand{\ineq}[1]{\footnotesize$#1$\normalsize}{}
\newcommand{\prior}{\text{{PYCARL}}}{}
\newcommand{\tech}{\text{{RENEU}}}{}

\begin{document}
\bstctlcite{IEEEexample:BSTcontrol}
%
\title{Improving Dependability of Neuromorphic Computing With Non-Volatile Memory}

\author{\IEEEauthorblockN{Shihao Song}
\IEEEauthorblockA{Drexel University\\
Philadelphia, PA 19104\\
Email: shihao.song@drexel.edu}
\and
\IEEEauthorblockN{Anup Das}
\IEEEauthorblockA{Drexel University\\
Philadelphia, PA 19104\\
Email: anup.das@drexel.edu}
\and
\IEEEauthorblockN{Nagarajan Kandasamy}
\IEEEauthorblockA{Drexel University\\
Philadelphia, PA 19104\\
Email: nk78@drexel.edu}
}

\maketitle

\begin{abstract}
\input{sections/abstract}
\end{abstract}


%
\IEEEpeerreviewmaketitle

\section{Introduction}\label{sec:introduction}
\input{sections/introduction}

\section{Background}\label{sec:background}
\input{sections/background}

\section{Reliability Formulation}\label{sec:formulation}
\input{sections/problem}

\section{Proposed Solution}\label{sec:approach}
\input{sections/approach}

\section{Results and Discussions}\label{sec:evaluation}
\input{sections/evaluation}

\section{Conclusion}\label{sec:conclusion}
\input{sections/conclusion}

\section*{Acknowledgment}
This work is supported by the National Science Foundation Faculty Early Career Development Award CCF-1942697 (CAREER: Facilitating Dependable Neuromorphic Computing: Vision, Architecture, and Impact on Programmability).



\bibliographystyle{IEEEtran}
\bibliography{IEEEabrv,edcc}
%



\end{document}

%% file: sections/abstract.tex
As process technology continues to scale aggressively, circuit aging in a neuromorphic hardware due to negative bias temperature instability (NBTI) and time-dependent dielectric breakdown (TDDB) is becoming a critical reliability issue and is expected to proliferate when using non-volatile memory (NVM) for synaptic storage. This is because an NVM requires high voltage and current to access its synaptic weight, which further accelerates the circuit aging in a neuromorphic hardware. Current methods for qualifying reliability are overly conservative, since they estimate circuit aging considering worst-case operating conditions and unnecessarily constrain performance. This paper proposes \tech{}, a reliability-oriented approach to map machine learning applications to neuromorphic hardware, with the aim of improving system-wide reliability without compromising key performance metrics such as execution time of these applications on the hardware. 
Fundamental to \tech{} is a novel formulation of the aging of CMOS-based circuits in a neuromorphic hardware considering different failure mechanisms. Using this formulation, \tech{} develops a system-wide reliability model which can be used inside a design-space exploration framework involving the mapping of neurons and synapses to the hardware. To this end, \tech{} uses an instance of Particle Swarm Optimization (PSO) to generate mappings that are Pareto-optimal in terms of performance and reliability.
We evaluate \tech{} using different machine learning applications on a state-of-the-art neuromorphic  hardware with NVM synapses. Our results demonstrate an average 38\% reduction in circuit aging, leading to an average 18\% improvement in the lifetime of the hardware compared to current practices. \tech{} only introduces a marginal performance overhead of 5\% compared to a performance-oriented state-of-the-art.

%% file: sections/introduction.tex
Machine learning models built with spike-based computations and bio-inspired learning algorithm, e.g., Spiking Neural Network (SNN)~\cite{maass1997networks} lower the energy consumption of machine learning tasks when they are executed on event-driven neuromorphic hardware such as DYNAP-SE~\cite{Moradi_etal18}, TrueNorth~\cite{debole2019truenorth}, and Loihi~\cite{davies2018loihi}. 
A typical neuromorphic hardware consists of computation units called \textit{neurosynaptic cores}, communicating spikes via a shared interconnect. A neurosynaptic core is essentially a crossbar, which is a \ineq{n\times n} organization of input and output neurons and synaptic weight storage at each crosspoint. 

Recently, Non-Volatile Memory (NVM) such as Phase Change Memory (PCM), Oxide-based RAM (OxRAM), and Spin-Transfer Torque Magnetic RAM (STT-MRAM) are used for synapses in neuromorphic architectures~\cite{burr2017neuromorphic}, beyond their use as memory for conventional von-Neumann computing \cite{songISMMb,songISMMa,song2019enabling}. NVMs bring certain advantages such as high integration density, multi-bit synapse, CMOS compatibility, and above all, non-volatility, which can further lower the energy consumption of neuromorphic hardware. However, NVMs also introduce reliability issues such as endurance, aging, and read disturbances (see Table~\ref{tab:reliability_summary}) \cite{chen2018reliability,gleixner2009reliability,pirovano2004reliability}. These issues are triggered when propagating current through an NVM synapse. In this work, we focus on one specific reliability issue --- that of aging of circuit components in a neuromorphic hardware leading to hard failures, and propose an intelligent solution to mitigate this problem (Section~\ref{sec:approach}). 

Although reliability issues of NVMs have been addressed at the system-level for von-Neumann computing with NVM-based main memory, e.g., \cite{jiang2014low,songISMMb}, these approaches do not apply to neuromorphic computing.
Recent system-level works on mapping SNN-based applications to neuromorphic hardware, such as~\cite{balaji20pycarl,songlctes2020,balaji2019mapping,ji2018bridge,das2018mapping,das2018dataflow,balaji2019framework2}, target performance improvement only. They do not consider reliability issues of neuromorphic computing. Our recent work~\cite{balaji2019framework} demonstrates the significant reliability degradation 
introduced using these SNN mapping approaches. In fact, this motivating work has also shown the change in reliability profile for different neuron and synapse mapping strategies. This is because with change in mapping, computation units in the hardware are stressed differently when processing a machine learning task, resulting in different reliability behavior of the hardware.

\textbf{Contributions:} This paper addresses reliability issues, specifically, circuit aging in neuromorphic computing with NVM using an intelligent neuron and synapse mapping technique. Following are our key contributions.

\begin{itemize}
    \item We formulate the detailed aging of CMOS-based circuits in a neuromorphic hardware. 
    
    \item We incorporate and integrate different failure mechanisms such as Time-Dependent Dielectric Breakdown (TDDB), Negative-Bias Temperature Instability (NBTI), and Hot Carrier Injection (HCI).
    
    \item We use this low-level circuit aging formulation to develop a system-wide reliability model, which allows to estimate the aging of computation units in a neuromorphic hardware when processing a spike train from a machine learning application.
    
    
    \item We propose a meta-heuristic approach based on Particle Swarm Optimization (PSO)~\cite{eberhart1995new} to map neurons and synapses to a neuromorphic hardware. We apply Pareto Optimization to retain only those mappings that are optimal in terms of reliability and performance.  
\end{itemize}

We evaluate our approach, which we call \textbf{\tech{}} (\underline{RE}liability-aware \underline{NEU}romorphic Computing), with state-of-the-art machine learning applications built using multi-layer perceptron (MLP), convolutional neural network (CNN), and recurrent neural network (RNN) models on a state-of-the-art neuromorphic hardware with OxRAM synapses. Results demonstrate an average of 38\% reduction in circuit aging, leading to an average of 18\% improvement in the lifetime of the hardware compared to current practices. \tech{} only introduces a marginal performance overhead of 5\% compared to a performance-oriented state-of-the-art mapping approach.

To the best of our knowledge, this is the first work that formulates the detailed aging of a neuromorphic hardware when executing a machine learning application and proposes a novel neuron and synapse mapping technique to reduce the overall aging of the hardware, improving dependability of neuromorphic computing.

%% file: sections/background.tex
\subsection{Spiking Neural Networks}
Spiking neural networks (SNNs) are computation models with spiking neurons and synapses~\cite{maass1997networks}.  
Neurons are typically implemented using Leaky Integrate-and-Fire (LIF) model~\cite{chicca2003vlsi}.
Information is represented using short impulses of infinitesimally small duration, called {spikes}. 
Spiking LIF neurons can be organized into feedforward layers, e.g., multi-layer perceptron (MLP) and convolutional neural network (CNN) or in a recurrent topologies, e.g., recurrent neural network (RNN). In this work, we evaluate MLP, CNN, and RNN-based machine learning applications (Section~\ref{sec:evaluation}). SNNs can implement both supervised and unsupervised learning. For supervised learning, a model is trained with examples from the field, without being exclusively programmed with any task-specific rules. A trained SNN model is then deployed on a neuromorphic hardware to perform inference from in-field data. For unsupervised learning, a machine learning model is trained in real-time using bio-inspired learning algorithms such as spike-timing dependent plasticity (STDP)~\cite{dan2004spike}. Without loss of generality, we focus on supervised machine learning approaches.

Recently, machine learning applications using analog computation models have achieved significant breakthroughs in computer vision and image processing domains~\cite{lecun2015deep}. Many research centers around the world now use analog models of CNNs for diverse applications. Section~\ref{sec:approach} discusses how \tech{} applies to analog CNNs.

\subsection{Neuromorphic Hardware}
\begin{figure}[h!]
	\centering
	\vspace{-5pt}
	\centerline{\includegraphics[width=0.99\columnwidth]{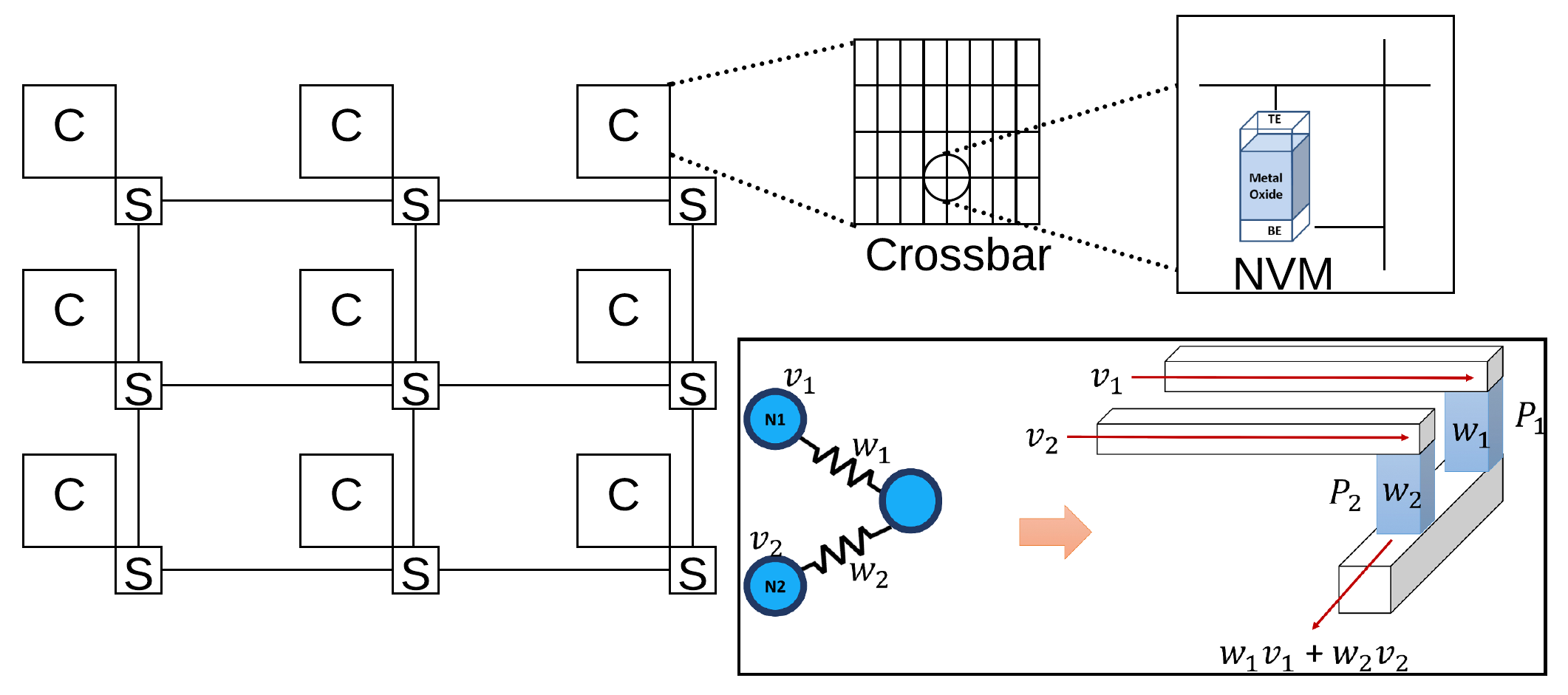}}
	\vspace{-5pt}
	\caption{A representative tile-based neuromorphic hardware.}
	\vspace{-10pt}
	\label{fig:neuromorphic_hardware}
\end{figure}

Figure~\ref{fig:neuromorphic_hardware} shows a representative neuromorphic hardware similar in structure to the DYNAP-SE, Loihi, and TrueNorth architectures. DYNAP-SE has four tiles per chip where each tile consists of a crossbar (C) and communicates with other tiles using the interconnect. Routing of spikes on the interconnect is facilitated using a switch (S).

A crossbar is an \ineq{n\times n}-organization (in 3D) of \ineq{n} rows (top electrodes) and \ineq{n} columns (bottom electrodes), and storage elements, e.g., NVM at their crosspoints. In DYNAP-SE, \ineq{n = 128}, while in TrueNorth, \ineq{n = 256}. When mapping an SNN to a neuromorphic hardware, synaptic weights are programmed as conductivity of these NVMs. The figure also illustrates a small example of mapping an SNN to the crossbar. 
Synaptic weights \ineq{w_{1}} and \ineq{w_{2}} are programmed into NVM cells P1 and  P2, respectively. The output spike voltages, \ineq{v_1} from N1 and \ineq{v_2} from N2, inject current into the crossbar, which is obtained by multiplying a pre-synaptic neuron's output spike voltage with the NVM cell's conductance at the cross-point of the pre- and post-synaptic neurons (following Ohm's law). Current summations along columns are performed in parallel using Kirchhoff’s current law, and implement the sums $\sum_j w_{i}v_i$, needed for forward propagation of neuron excitation.

A crossbar introduces the following \textbf{constraints} when mapping SNNs to a neuromorphic hardware.
\begin{itemize}
    \item Each crossbar has \ineq{n} input ports and \ineq{n} output ports, i.e., \ineq{n} input neurons, one at each row, inject current into the crossbar, and \ineq{n} output neurons, one at each column, act as current sink to propagate neuron computations.
    
    \item Each crossbar can accommodate a maximum of \ineq{n} pre-synaptic connections per output neuron.
\end{itemize}


\subsection{Non-volatile Memory}\label{sec:bacl_pcm}
Emerging NVM technologies such as phase-change memory (PCM), oxide-based memory (OxRAM), spin-based magnetic memory (STT-MRAM), and Flash have recently been used as synaptic storage elements within crossbars. NVMs are non-volatile, have high CMOS compatibility, and can achieve high integration density. Each NVM device can implement both a single-bit and multi-bit synapse. 
Because of these properties, an NVM-based neuromorphic hardware typically consumes energy that is magnitudes lower than using SRAMs~\cite{mallik2017design,garbin2015hfo,burr2017neuromorphic,ramasubramanian2014spindle}. However, NVMs also introduce reliability issues and Table~\ref{tab:reliability_summary} summarizes the sources of reliability concerns. 

\begin{table}[h]
\renewcommand{\arraystretch}{1.5}
\setlength{\tabcolsep}{2pt}
\caption{Reliability issues in NVMs.}
\label{tab:reliability_summary}
\centering
{\fontsize{10}{10}\selectfont
\begin{tabular}{|l|c|}
\hline
\textbf{Reliability Issues} & \textbf{NVMs}\\
\hline
High-voltage related circuit aging & PCM, Flash\\
High-current related circuit aging & OxRAM, STT-MRAM\\
Read disturbance & All\\
Limited endurance & All\\
\hline
\end{tabular}}
\end{table}

\begin{figure}[h!]
	\centering
	\vspace{-5pt}
	\centerline{\includegraphics[width=0.99\columnwidth]{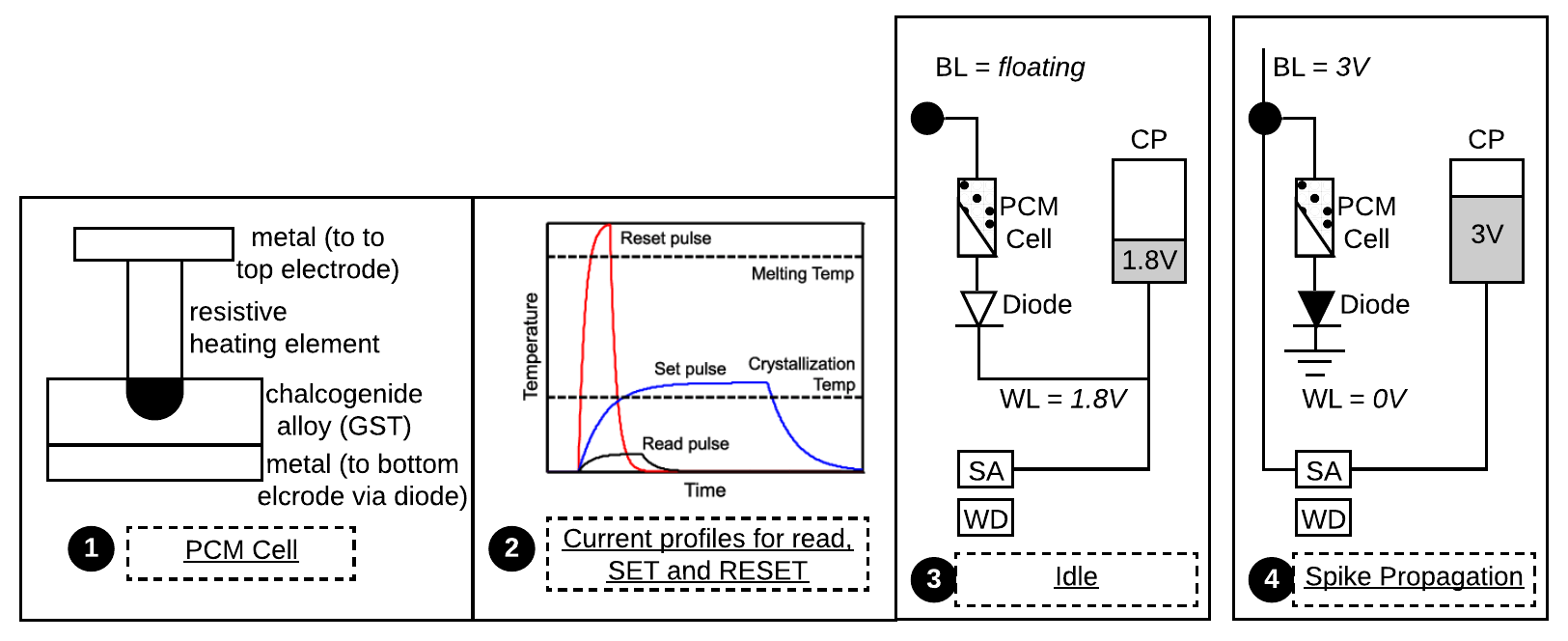}}
	\caption{Operation of PCM in neuromorphic computing.}
	\vspace{-10pt}
	\label{fig:pcm}
\end{figure}

In this work, we focus on PCM-based neuromorphic computing~\cite{burr2017neuromorphic}. Figure~\ref{fig:pcm}~\ding{182} illustrates how a chalcogenide semiconductor alloy is used to build a PCM cell.
The amorphous phase (RESET) in this alloy has higher resistance than the crystalline phase (SET).
Ge${}_2$Sb${}_2$Te${}_5$ (GST) is the most commonly used alloy for PCM.
To compute \ineq{(x_i\cdot w_i)}, a current is injected into the resistor-chalcogenide junction via the heater element. The current is controlled to ensure that the phase of the PCM cell is not disturbed. This is the fundamental operation of forward propagation of neuron excitation during inference. For online learning (e.g., using STDP), the injected current induces (Joule) heating in the chalcogenide alloy, changing its conductivity, thereby achieving synaptic weight updates. 
Figure~\ref{fig:pcm}~\ding{183} illustrates the current profiles necessary for inference (using the read pulse) and online learning (using the SET and RESET pulse) in PCM. 
These current profiles are generated using an on-chip charge pump (CP).
Figure~\ref{fig:pcm}~\ding{184} illustrates the PCM cell's operation when idle, i.e., when a neuron is not activated. We illustrate a 1D-1R structure, where a single PCM cell is connected to a row and column using a diode as an access device. Diode-based PCM cells allow very high integration density in scaled technology nodes compared to transistor-based PCM. The CP is operated at 1.8V to maintain the required biasing. Finally, Figure~\ref{fig:pcm}~\ding{185} illustrates the PCM operation during inference. The CP is operated at 3V to generate the read current profile of Figure~\ref{fig:pcm} \ding{183} using the sense amplifier (SA). The write driver (WD) is used for generating the currents for online learning.

These high-voltage operations of the charge pump (and the peripheral circuit of a crossbar) accelerates circuit aging, lowering the dependability of neuromorphic computing. Section~\ref{sec:formulation} formulates aging in neuromorphic computing and Section~\ref{sec:approach} proposes our solution to this problem.

\textbf{Dependability issues in transistor-based PCM cells:} When using transistor-based PCM cells, the CP is operated at lower voltages: 1.2V during idle and 1.8V during spike propagation. Though aging issues are less severe in such designs, they are still dependability concerns for neuromorphic computing. Our solution, which we describe in Section~\ref{sec:approach}, applies to both diode and transistor-based PCM cells, and improves dependability for both (see Section~\ref{sec:res_diode_bjt}).

%% file: sections/problem.tex
\begin{figure}[h!]
	\centering
	\vspace{-5pt}
	\centerline{\includegraphics[width=0.99\columnwidth]{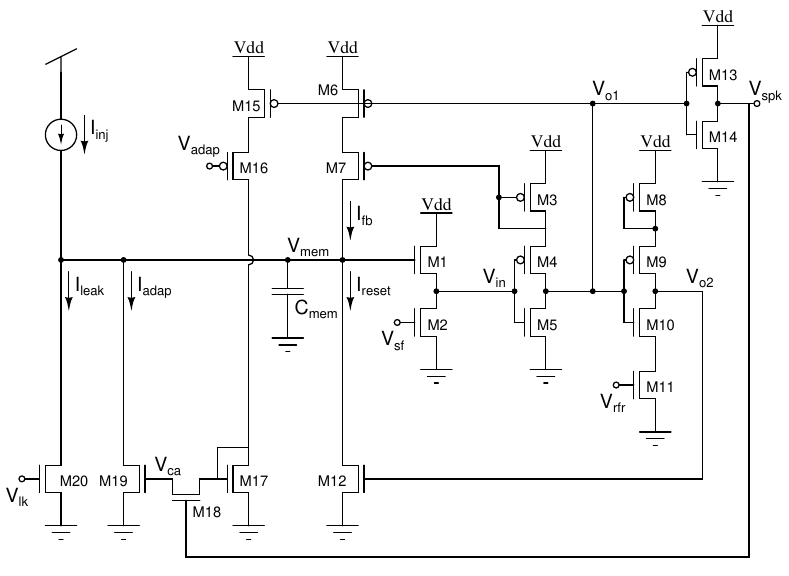}}
	\caption{Internal architecture of a neuron.}
	\vspace{-10pt}
	\label{fig:neuron}
\end{figure}

There are many sources of reliability issues in a neuromorphic hardware with PCM synapses, as listed in Table~\ref{tab:reliability_summary}. We focus on the aging of CMOS-based circuits due to high voltage PCM operations. Figure~\ref{fig:neuron} shows the internal circuitry of a neuron which injects current into a crossbar~\cite{indiveri2003low}. We observe that the CMOS transistors are operated at elevated voltages (1.8V and 3V for 1D-1R PCM, and 1.2V and 1.8V for 1T-1R PCM) during the execution of a machine learning application. These elevated voltages accelerate CMOS aging, leading to hard or soft faults in the neuromorphic hardware.
It is important to note that continuous device scaling and elevated operating temperatures can make these errors manifest sooner than endurance-related failures, making CMOS aging a critical dependability issue of NVM-based neuromorphic computing.

\subsection{High-Voltage Related Circuit Aging}
In this section we formulate CMOS aging considering 
Time-Dependent Dielectric Breakdown (TDDB), Negative-Bias Temperature Instability (NBTI), and Hot-Carrier Injection (HCI) failure mechanims. These are the dominant ones in scaled technology nodes (45nm and below). In older nodes, Electromigration (EM) plays a key role~\cite{das2015execution,das2014energy,das2013aging,SrinivasanISCA04}. Nevertheless, our aging formulation can be easily extended to also consider EM and any other failure mechanisms.

CMOS aging is accelerated when the device is \emph{stressed}, i.e., exposed to high overdrive voltages\footnote{Overdrive voltage is defined as the voltage between transistor gate and source ($V_{GS}$) in excess of the threshold voltage ($V_\text{th}$), where $V_\text{th}$ is the minimum voltage required between gate and source to turn the transistor on.}. With this understanding, 
we provide a brief background of these failure mechanisms. 
\begin{itemize}
    \item \textit{TDDB:} This is a failure mechanism in a CMOS device, when the gate oxide breaks down as a result of long-time application of relatively low electric field (as opposed to immediate breakdown, which is caused by strong electric field)~\cite{roussel2018new}. The lifetime of a CMOS device is measured in terms of its \textit{mean time to failure} (\textbf{MTTF}) as
\vspace{-6pt}
\begin{equation}
    \label{eq:MTTF_TDDB}
    \vspace{-6pt}
    \footnotesize \text{MTTF}_\text{TDDB} = A.e^{-\gamma\sqrt{V}},
\end{equation}
where \ineq{A} and \ineq{\gamma} are material-related constants, and \ineq{V} is the overdrive gate voltage of the CMOS device. 
\item \emph{NBTI:} This is a failure mechanism in a CMOS device in which positive charges are trapped at the oxide-semiconductor boundary underneath the gate~\cite{gao2017nbti}.
NBTI manifests as 1) decrease in drain current and transconductance, and 2) increase in off current and threshold voltage. 
The NBTI lifetime of a CMOS device is
\vspace{-6pt}
\begin{equation}
    \label{eq:MTTF_NBTI}
    \vspace{-6pt}
    \footnotesize \text{MTTF}_\text{NBTI} = \frac{A}{V^\gamma}e^{\frac{E_a}{KT}},
\end{equation}
where \ineq{A} and \ineq{\gamma} are material-related constants, \ineq{E_a} is the activation energy, \ineq{K} is the Boltzmann constant, \ineq{T} is the temperature, and \ineq{V} is the overdrive gate voltage of the CMOS device.
\item \emph{HCI:} This is a failure mechanism in a CMOS device, when a carrier (electron or hole) gains sufficient kinetic energy to overcome the potential barrier of the conducting channel and gets trapped in the gate dielectric, permanently changing the switching characteristics of the device~\cite{wan2019hci}.
\end{itemize}

Unlike the TDDB and NBTI failure mechanisms, for which silicon-characterized reliability models are available from foundries, characterized models for HCI failure mechanism are still in development for scaled nodes.

\subsubsection{\underline{TDDB Aging of a Single Neuron in a Tile}}

We illustrate our aging formulation for TDDB failure mechanism first, and then show how to extend this formulation to consider other failure mechanisms such as NBTI and HCI. 

TDDB failures can also be modeled using the Weibull distribution~\cite{fenner2018making} with a scale parameter \ineq{\alpha} and a slope parameter \ineq{\beta}. Reliability at time \ineq{t} can be written as
\begin{equation}
\label{eq:eq1}
\scriptsize R(t) = e^{-\left(\frac{t}{\alpha(V)}\right)^\beta},
\end{equation}
with the corresponding MTTF computed as 
\begin{equation}
\label{eq:2}
\scriptsize MTTF = \int_{0}^{\infty}R(t)dt = \alpha(V)\Gamma\left(1+ \frac{1}{\beta}\right),
\end{equation}
where \ineq{\Gamma} is the Gamma function. Using the expressions for MTTF from Equations~\ref{eq:MTTF_TDDB} and \ref{eq:eq1}, and rearranging, we obtain the expression for the scale parameter \ineq{\alpha} as
\begin{equation}
\label{eq:eq3}
\scriptsize \alpha(V) = \frac{A.e^{-\gamma\sqrt{V}}}{\Gamma\left(1+\frac{1}{\beta}\right)}.
\end{equation}

\begin{figure}[h!]
	\centering
	\vspace{-5pt}
	\centerline{\includegraphics[width=0.99\columnwidth]{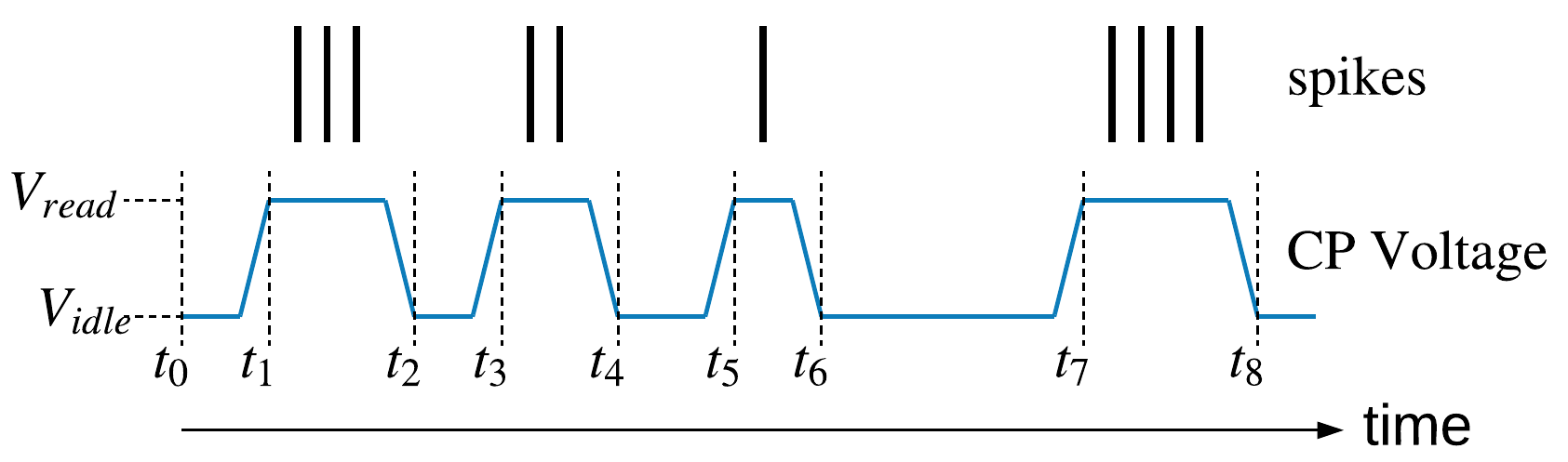}}
	\caption{Operating voltage of the neuron circuit to propagate spikes through the crossbar.}
	\vspace{-10pt}
	\label{fig:reliability_demo}
\end{figure}

Figure~\ref{fig:reliability_demo} illustrates a spike train and the change in operating voltage of a neuron circuit to inject current into the crossbar for each spike in the spike train.
To estimate the aging in this time duration, we let \ineq{[t_i,t_{i+1})} be the \ineq{(i+1)}\ineq{{}^\text{th}} time interval with \ineq{\Delta t_i = t_{i+1} - t_i} and \ineq{V_i} as the neuron's voltage.

Reliability of a CMOS device at the start of execution is
\begin{equation}
\label{eq:eq4}
\scriptsize R(t)|_{t=t_0} = 1.
\end{equation}
At the end of the first interval, the reliability is
\begin{equation}
\label{eq:eq5}
\scriptsize R(t_1^-) = e^{-\left(\frac{t_1}{\alpha(V_0)}\right)^\beta}.
\end{equation}
Using the term $\theta$ to represent reliability degradation during this interval \ineq{[t_o,t_1)}, the reliability at the beginning of the second interval (i.e., right after the start of the first idle period) is
\begin{equation}
\label{eq:eq7}
\scriptsize R(t_1^+) = e^{-\left(\frac{t_1+\theta}{\alpha(V_1)}\right)^\beta}.
\end{equation}
Due to the continuity of the reliability function, we can equate Equations~\ref{eq:eq5} \& \ref{eq:eq7} to compute $\theta$ as
\begin{equation}
\label{eq:eq8}
\scriptsize \theta = \left(\frac{\alpha(V_1)}{\alpha(V_0)} - 1\right)t_1.
\end{equation}
Substituting Eq.~\ref{eq:eq8} in Eq.~\ref{eq:eq7}, reliability at time $t_2$ is 
\begin{equation}
\label{eq:eq9}
\scriptsize R(t_2) = e^{-\left(\frac{\Delta t_1}{\alpha(V_1)} + \frac{\Delta t_0}{\alpha(V_0)}\right)^{\beta}}.
\end{equation}
We can extend this equation to compute the reliability of a CMOS device in the neuron circuit at the end of the spike train as
\begin{equation}
\label{eq:eq10}
\scriptsize R(t_s) = e^{-\left(\sum_{i=1}^{n}\frac{\Delta t_i}{\alpha(V_i)}\right)^{\beta}}.
\end{equation}
We define the TDDB \emph{aging} of the neuron \ineq{i}, \ineq{\mathcal{A}_\text{TDDB}(i)}, as
\begin{equation}
\label{eq:eq11_pre1}
\scriptsize \mathcal{A}_\text{TDDB}(i) = \sum_{i=1}^{n}\frac{\Delta t_i}{\alpha(V_i)}, \text{ such that } R(t_s) = e^{-\left(\mathcal{A}_\text{TDDB}(i)\right)^{\beta}},
\end{equation}
where the scaling factor \ineq{\alpha(V_i)} can be calculated using Eq. \ref{eq:eq3}.

\subsubsection{\underline{TDDB Aging of a Tile}}
The aging of Eq.~\ref{eq:eq11_pre1} can be extended to incorporate the aging of other neurons in the tile (i.e., all input to a crossbar in the tile). To this end, we consider a tile to be faulty if any input neuron fails due to TDDB.\footnote{In our future work, we consider a single faulty neuron to reduce the capacity of the crossbar in a tile, rather than treating the entire tile to be faulty. Nonetheless, considering this capacity reduction is a reactive solution. The proposed work, on the other hand is an orthogonal proactive approach.} Using this series failure model, the TDDB aging of tile~\ineq{j} is 
\begin{equation}
    \label{eq:overall_tddb_aging}
    \footnotesize \mathcal{A}^j_\text{TDDB} = \max\{\mathcal{A}^j_\text{TDDB}(i)\}, \forall i\in 1,\cdots,n.
\end{equation}
where \ineq{n} is the number of input ports of the tile, i.e., the number of input neurons and \ineq{\mathcal{A}^j_\text{TDDB}(i)} is the TDDB aging of the \ineq{i^\text{th}} neuron in the \ineq{j^\text{th}} tile, computed using Eq.~\ref{eq:eq11_pre1}.

\subsubsection{\underline{TDDB Aging of a Neuromorphic Hardware}} To compute the TDDB aging of the neuromorphic hardware with \ineq{N} tiles, we similarly use a series failure model, where a single faulty tile leads to the failure of the neuromorphic hardware. The TDDB aging of the neuromorphic hardware is 
\begin{equation}
    \label{eq:overall_neuromorphic}
    \footnotesize \mathcal{A}_\text{TDDB} = \max\{\mathcal{A}^j_\text{TDDB}\}, \forall j\in 1,\cdots,N.
\end{equation}

Formulating TDDB as a series failure model allows our mapping framework to minimize the maximum aging, i.e., solving a minmax problem (see Section~\ref{sec:approach}).

\subsection{{Combining Aging due to other Failure Mechanisms}} 
Next, we consider 
NBTI, which is manifested as threshold voltage shift in a CMOS device. 
Recent works such as~\cite{gao2017nbti} suggest that NBTI is the collective response of two independent mechanisms: the \textit{as-grown hole traps} (AHTs) and \textit{generated defects} (GDs). AHTs and a small proportion of GDs can be recovered by annealing at high temperatures if the NBTI stress voltage is removed. 

\begin{figure}[h!]
 	\centering
    \vspace{-6pt}
 	\centerline{\includegraphics[width=0.5\columnwidth]{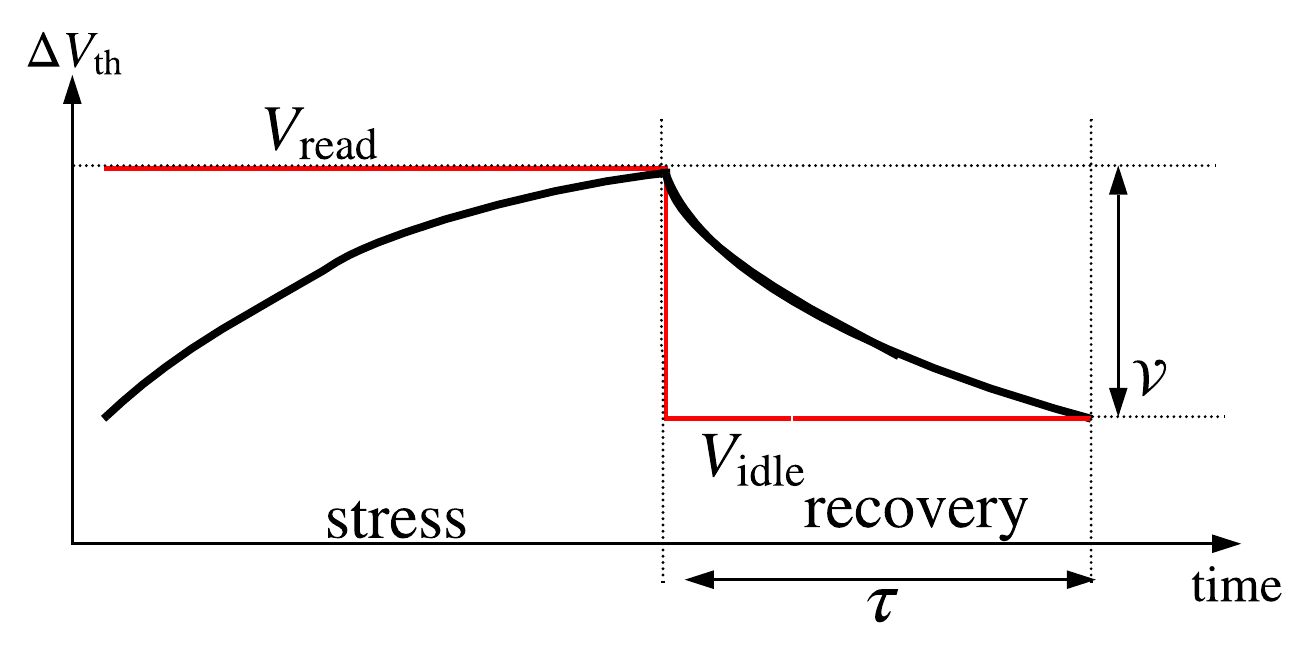}}
 	\caption{Demonstration of degradation due to NBTI.}
    \vspace{-10pt}
 	\label{fig:nbti_demo}
\end{figure}

Figure~\ref{fig:nbti_demo} illustrates the stress and recovery of threshold voltage of a CMOS device due to NBTI failure mechanism on application of a high (\ineq{V_\text{read}}) and a low voltage (\ineq{V_\text{idle}}). We observe that both stress and recovery depends on the time of exposure to the corresponding voltage level. This implies that when a neuron is idle, the NBTI aging of the neuron recovers from stress. This is different from TDDB mechanism, where the neuron continues to undergo TDDB aging even when idle.

Using the spike train of Fig.~\ref{fig:reliability_demo}, NBTI aging is given by 
\begin{equation}
\label{eq:eq10_nbti}
\scriptsize \mathcal{A}_\text{NBTI} = \sum_{i=0}^{m-1}g_0\cdot(V_i-V_\text{NBTI})^m\cdot(t_{i+1}-t_i)^n, \text{ such that } R(T) = e^{-\mathcal{A}_\text{NBTI}^{\beta}},
\end{equation}
where \ineq{\beta,g_0,m,n} are material-dependent constants~\cite{gao2017nbti}.

To combine the aging from different failure mechanisms such as TDDB, NBTI and HCI, we use the Sum-of-Failure-Rates (SOFR) model, which is used extensively in the industry~\cite{amari1997closed}. SOFR assumes an exponential lifetime distribution for each failure mechanism, allowing us to compute the overall aging of the neuron as the combined effect of aging due to each failure mechanism individually.

Using Equations~\ref{eq:2} \& \ref{eq:eq11_pre1}, the failure rate for TDDB is 
\begin{equation}
    \label{eq:failure_rate}
    \footnotesize \lambda_\text{TDDB} = \frac{1}{\text{MTTF}_\text{TDDB}} = \int_0^\infty e^{\left(\mathcal{A}_\text{TDDB}\right)^\beta}.
\end{equation}

Using SOFR, the overall failure rate is computed as
\begin{equation}
    \label{eq:overall_failure_rate}
    \footnotesize \lambda_\text{overall} = \frac{1}{\text{MTTF}_\text{overall}} = \lambda_\text{TDDB} + \lambda_\text{NBTI} +  \lambda_\text{HCI}.
\end{equation}

The overall aging of the neuromorphic hardware is therefore
\begin{equation}
    \label{eq:overall_aging}
    \footnotesize \mathcal{A}_\text{overall} = \ln{\left(e^{\left(\mathcal{A}_\text{TDDB}\right)^\beta + } + e^{\left(\mathcal{A}_\text{NBTI}\right)^\beta + } + e^{\left(\mathcal{A}_\text{HCI}\right)^\beta + }\right)}^\frac{1}{\beta}.
\end{equation}

Equation~\ref{eq:overall_aging} can be extended to consider other failure mechanisms, as well as other models to compute the aging. 

\subsection{Lifetime Computation}
The \emph{lifetime} of a neuromorphic hardware is usually \emph{much longer} than the {execution time} of a machine learning workload. 
To estimate {how much} a neuron circuit's lifetime changes due to the mapping of neurons and synapses to the hardware, we {estimate} the {aging} over the time duration of a workload, and then use it to {extrapolate} the lifetime, considering the {same} workload being executed {repeatedly} until one of the hardware components fails due to one of the failure mechanism. The lifetime measured as MTTF is
\begin{equation}
    \label{eq:MTTF_final_final}
    \footnotesize MTTF = \int_0^\infty e^{\left(\mathcal{A}_\text{overall}\right)^-\beta}
\end{equation}

To compute MTTF, the slope parameter of Weibull distribution is set to \ineq{\beta = 2}, and the operating temperature is set to \ineq{300K}. Other fitting parameters are adjusted to achieve an MTTF of \emph{2 years} in the baseline system. This is the typical lifetime of neuromorphic products. 

%% file: sections/approach.tex
\subsection{High-Level Overview}
\begin{figure}[h!]
	\centering
	\vspace{-5pt}
	\centerline{\includegraphics[width=0.99\columnwidth]{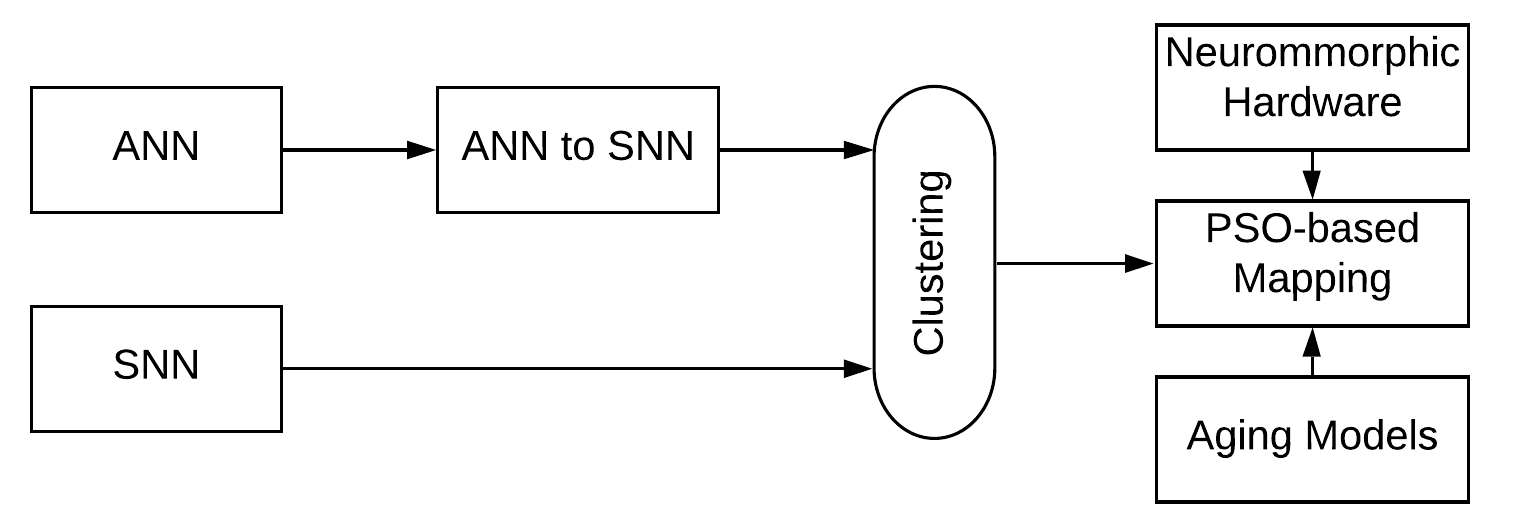}}
	\caption{Workflow of \tech{}.}
	\vspace{-10pt}
	\label{fig:overview}
\end{figure}

Figure~\ref{fig:overview} shows a high-level overview of the workflow of \tech{}, which can either directly input an SNN-based machine learning application or an artificial neural network (ANN)-based one after converting its ANN operations to SNN using the N2D2 tool~\cite{bichler2017n2d2,balaji2018power}. The SNN model is then partitioned into clusters, where each cluster accommodates a fixed number of neurons and synapses, and can fit on its entirety to a crossbar in the hardware. We use the clustering technique of the DFSynthesizer tool~\cite{songlctes2020}. The clustered SNN model is then used by \tech{} to find the mapping of the clusters to the crossbars using an instance of the PSO. We now describe in details the PSO step of \tech{}.

\subsection{PSO-based Cluster Mapping}
We consider the mapping of a clustered SNN \ineq{G(C,E)} with a set \ineq{C} of clusters and a set \ineq{E} of edges, to the neuromorphic hardware \ineq{H(T,I)}, where \ineq{T} is the set of tiles in the hardware and \ineq{I} is the set of connections of these tiles.

Mapping \ineq{M:{G(C,E)}\rightarrow {H(T,I)}} is specified by a logical matrix \ineq{(m_{ij}) \in \{0,1\}^{|\mathcal{C}|\times|\mathcal{T}|}}, where \ineq{m_{ij}} is defined as
\begin{equation}
    \label{eq:mapping_rep}
    \footnotesize m_{ij} = \begin{cases}
    1 & \text{if cluster } {c}_i \in \mathcal{C} \text{ is mapped to tile } {T}_j\in T\\
    0 & \text{otherwise}
    \end{cases}
\end{equation}

The mapping constraints are the following:
\begin{itemize}
    \item[1.] A cluster can be mapped to only one crossbar, i.e.,
    \begin{equation}
        \label{eq:mapping_constraint_1}
        \footnotesize \sum_j m_{ij} = 1~~~\forall i
    \end{equation}
    \item[2.] A crossbar can accommodate at most one cluster, i.e.,
    \begin{equation}
        \label{eq:mapping_constraint_2}
        \footnotesize \sum_i m_{ij} \leq 1~~~\forall j
    \end{equation}
\end{itemize}

We use an instance of Particle Swarm Optimization (PSO)~\cite{eberhart1995new} to obtain this mapping.
The fitness function of the PSO is a joint metric \ineq{\lambda} defined as the product of aging and execution time. This metric \ineq{\lambda} is computed as follows.
\begin{itemize}
    \item \ineq{\tau_M} = Execution time of mapping \ineq{M}, computed using the \texttt{DFSynthesizer} tool \cite{songlctes2020}.
    \item \ineq{\mathcal{A}_M} = Aging of mapping \ineq{M}, computed using (\ref{eq:overall_aging}) with spike trains obtained from the SNN model.
    \item \ineq{\lambda_M = \tau_M\cdot \mathcal{A}_M}. This is the fitness function.
\end{itemize}

The PSO finds the optimum mapping with the minimum value of the fitness function, i.e.,
\begin{equation}
    \label{eq:opt_obj}
    \footnotesize \lambda_{M_\text{opt}} = \min\{\lambda_{M_i} | i\in 1,2,\cdots\},
\end{equation}

We instantiate \ineq{n_p} swarm particles. The position of these particles are solutions to the fitness function, and they represent cluster mappings, i.e., \ineq{M}'s in Equation \ref{eq:opt_obj}.
Each particle also has a velocity with which it moves in the search space to find the optimum solution. During the movement, a particle updates its position and velocity according to its own experience (closeness to the optimum) and also experience of its neighbors. We introduce the following notations.
 
\begin{footnotesize}
 	\begin{align}
 	\label{eq:pso_defn}
 	D = |\mathcal{C}|\times|\mathcal{V}| &= \text{dimensions of the search space}\\
 	\mathbf{\Theta} = \{\mathbf{\theta}_l\in\mathbb{R}^{D}\}_{l=0}^{n_p-1} &= \text{positions of particles in the swarm}\nonumber \\
 	\mathbf{V} = \{\mathbf{v}_l\in\mathbb{R}^{D}\}_{l=0}^{n_p-1} &= \text{velocity of particles in the swarm}\nonumber 
 	\end{align}
 \end{footnotesize}
Position and velocity of swarm particles are updated, and the fitness function is computed as

\begin{footnotesize}
	\begin{align}
	\label{eq:pos_vel_update}
	\mathbf{\Theta}(t+1) &= \mathbf{\Theta}(t) + \mathbf{V}(t+1)\\
	\mathbf{V}(t+1) &= \mathbf{V}(t) + \varphi_1\cdot\Big(P_{\text{best}}-\mathbf{\Theta}(t)\Big) + \varphi_2\cdot\Big(G_{\text{best}}-\mathbf{\Theta}(t)\Big)\nonumber\\
	F(\theta_l) &= \lambda_{\theta_l} = \tau_{\theta_l}\cdot\mathcal{A}_{\theta_l}\nonumber
	\end{align}
\end{footnotesize}
\normalsize where $t$ is the iteration number, $\varphi_1,\varphi_2$ are constants and $P_{\text{best}}$ (and $G_{\text{best}}$) is the particles own (and neighbors) experience. 
Finally, local and global bests are updated as

\vspace{-5pt}
\begin{footnotesize}
    \begin{align}
    \label{eq:pbest}
        && P_\text{best}^l = F(\theta_l) \text{ if } F(\theta_l) < F(P_\text{best}^l)\nonumber \\
        && G_\text{best} = \displaystyle \min_{l=0,\dots n_p-1} P_\text{best}^l
    \end{align}
\end{footnotesize}

Due to the binary formulation of the mapping problem (see Equation \ref{eq:mapping_rep}), we need to binarize the velocity and position of Equation \ref{eq:pso_defn}, which we illustrate below.

\vspace{-5pt}
\begin{footnotesize}
	\begin{align}
	\label{eq:binarization}
	&\hat{\mathbf{V}} = \texttt{sigmoid}(\mathbf{V}) = \frac{1}{1+\texttt{e}^{-\mathbf{V}}} \nonumber \\
	&	\hat{\Theta} = \begin{cases}
	0 \text{~~if } \texttt{rand()} < \hat{\mathbf{V}}\\
	1 \text{~~otherwise }
	\end{cases}
	\end{align}
\end{footnotesize}


In finding a new position of a PSO particle, we use the two {constraints} (\ref{eq:mapping_constraint_1}) and (\ref{eq:mapping_constraint_2}).

\subsection{Pareto-Optimization}
\begin{figure}[h!]
	\centering
	\vspace{-5pt}
	\centerline{\includegraphics[width=0.99\columnwidth]{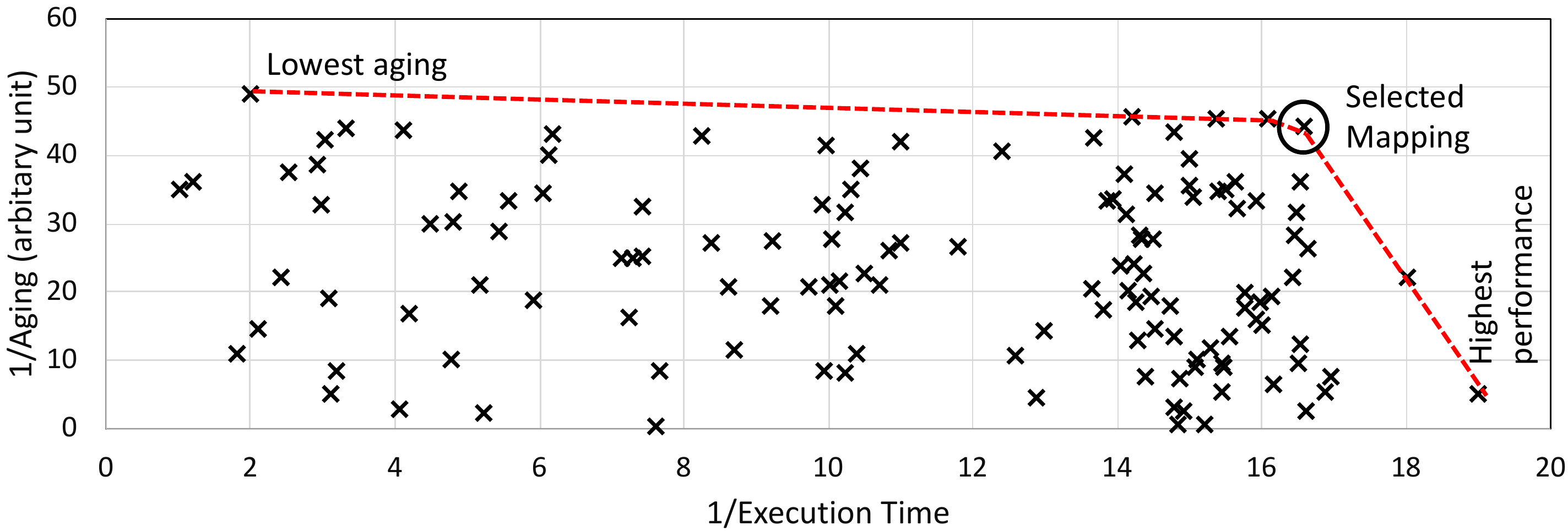}}
	\caption{Pareto optimization.}
	\vspace{-10pt}
	\label{fig:pareto}
\end{figure}

We record all mappings generated using the PSO.
In the final step, we perform Pareto-optimization to select a mapping that maximizes aging without compromising performance. Figure \ref{fig:pareto} shows the Pareto front of LeNet-CIFAR application and selection of the final mapping.

%% file: sections/evaluation.tex
We conduct all simulations on a system with 8 CPUs, 32GB RAM, and NVIDIA Tesla GPU, running Ubuntu 16.04. 
We model the DYNAP-SE neuromorphic hardware~\cite{Moradi_etal18} with four tiles. Each tile has one \ineq{128\times 128} crossbar with PCM synapses.
We evaluate 10 standard machine learning applications obtained from~\cite{balaji20pycarl} that are summarized in Table~\ref{tab:apps}.


\begin{table}[h!]
	\renewcommand{\arraystretch}{0.8}
	\setlength{\tabcolsep}{2pt}
	\centering
	\begin{threeparttable}
	{\fontsize{6}{10}\selectfont
		\begin{tabular}{cc|ccl}
			\hline
			\textbf{Class} & \textbf{Applications} & \textbf{Synapses} & \textbf{Neurons} & \textbf{Topology}\\
			\hline
			\multirow{3}{*}{MLP} & EdgeDet & 272,628 &  1,372 & FeedForward (4096, 1024, 1024, 1024)\\
			& ImgSmooth & 136,314 & 980 & FeedForward (4096, 1024)\\
			& MLP-MNIST & 79,400 & 984 & FeedForward (784, 100, 10)\\
			\hline
			\multirow{4}{*}{CNN} & CNN-MNIST & 159,553 & 5,576 & CNN\tnote{a}\\
			& LeNet-MNIST & 1,029,286 & 4,634 & CNN\tnote{b}\\
			& LeNet-CIFAR & 2,136,560 & 18,472 & CNN\tnote{c}\\
			& HeartClass~\cite{das2018heartbeat} & 2,396,521 & 24,732 & CNN\tnote{d}\\
			\hline
 			\multirow{3}{*}{RNN} & HeartEstm~\cite{das2018unsupervised} & 636,578 & 6,952 & Recurrent Reservoir\\
 			& SpeechRecog & 636,578 & 6,952 & Recurrent Reservoir\\
 			& VisualPursuit & 636,578 & 6,952 & Recurrent Reservoir\\
			\hline
	\end{tabular}}
	\begin{tablenotes}\scriptsize
        \item[a.] Input(24x24) - [Conv, Pool]*16 - FC*150 - FC*10
        \item[b.] Input(32x32) - [Conv, Pool]*6 - [Conv, Pool]*16 - Conv*120 - FC*84 - FC*10
        \item[c.] Input(32x32x3) - [Conv, Pool]*6 - [Conv, Pool]*6 - FC*84 - FC*10
        \item[d.] Input(82x82) - [Conv, Pool]*16 - [Conv, Pool]*16 - FC*256 - FC*6
    \end{tablenotes}
	\end{threeparttable}
	\vspace{12pt}
	\caption{Applications used to evaluate our approach.}
	\label{tab:apps}
\end{table}

We evaluate the following state-of-the-art approaches.
\begin{itemize}
    \item \emph{\underline{PYCARL}}: A performance-oriented approach to map neurons and synapses to a neuromorphic hardware~\cite{balaji20pycarl}.
    
    \item \emph{\underline{Reliability Qualification}}: A conservative reliability qualification technique, which estimates aging assuming worst-case operating conditions~\cite{balaji2019framework}.
    
    \item \emph{\underline{\tech{}}} (proposed): We use a detailed circuit aging model and use it to map neurons and synapses to a neuromorphic hardware improving reliability without compromising performance.
\end{itemize}

\subsection{Lifetime Improvement}
\begin{figure}[h!]
	\centering
	\vspace{-5pt}
	\centerline{\includegraphics[width=0.99\columnwidth]{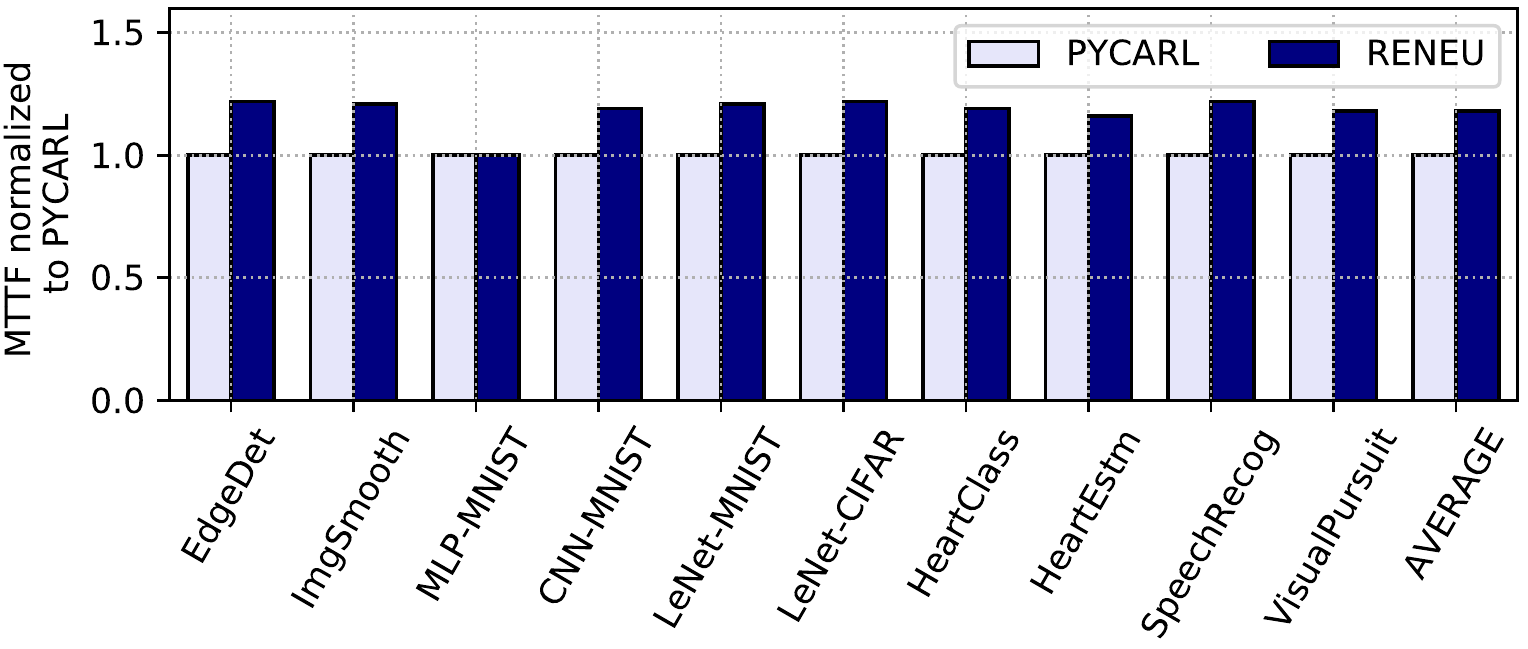}}
	\caption{MTTF normalized to \prior{} (higher is better).}
	\vspace{-10pt}
	\label{fig:mttf}
\end{figure}

Figure~\ref{fig:mttf} reports the MTTF of \tech{} normalized to \prior{} for each of our machine learning applications. 
We observe that the MTTF of \tech{} is better than \prior{} by an average of 18\%. This improvement is because \tech{} allocates clusters to tiles, minimizing the circuit aging of its crossbars. Lower aging leads to higher MTTF. We observe no noticeable improvement of MTTF for MLP-MNIST because this is a very small application to begin with.

\subsection{Aging Reduction}
\begin{figure}[h!]
	\centering
	\vspace{-5pt}
	\centerline{\includegraphics[width=0.99\columnwidth]{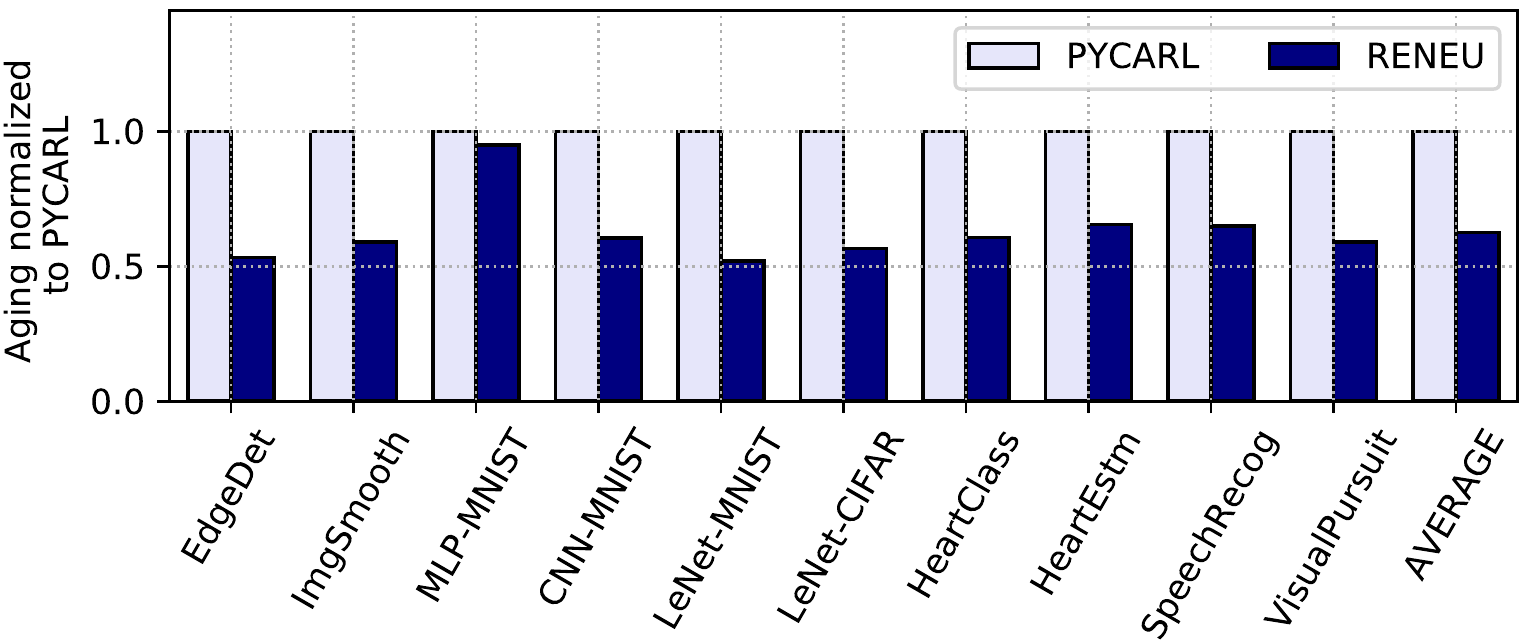}}
	\caption{Circuit aging in the neuromorphic hardware normalized to \prior{} (lower is better).}
	\vspace{-10pt}
	\label{fig:aging}
\end{figure}

Figure~\ref{fig:aging} reports the circuit aging caused by \tech{} normalized to \prior{} for each of our machine learning applications. 
We observe that the aging of \tech{} is lower than \prior{} by an average of 38\%. This improvement is because \tech{} formulates the detailed circuit aging of a neuromorphic hardware and allocates the neurons and synapses of a machine learning application to minimize it. 

\subsection{Temperature Dependency}
\begin{figure}[h!]
	\centering
	\vspace{-5pt}
	\centerline{\includegraphics[width=0.99\columnwidth]{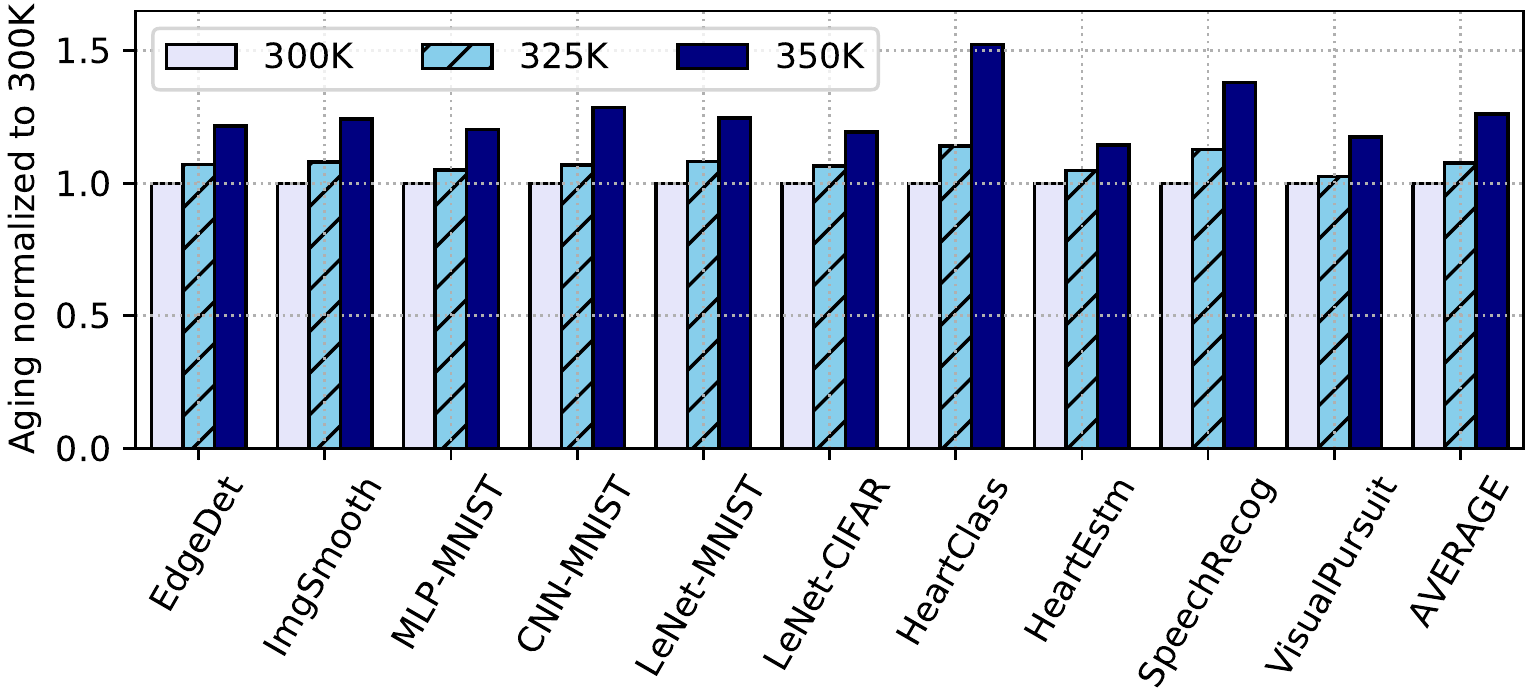}}
	\caption{Circuit aging of \tech{} at 325K and 350K normalized to \tech{} at 300K.}
	\vspace{-10pt}
	\label{fig:temperature}
\end{figure}

Figure~\ref{fig:temperature} illustrates the temperature dependency of circuit aging in a neuromorphic hardware. We report the aging results of \tech{} at two elevated temperatures, 325K and 350K, for each of our machine learning applications. Aging results are normalized to \tech{} at 300K.
We observe that aging increases with an increase in temperature. Aging observed at 325K and 350K is higher than that observed at 300K by an average of 7\% and 26\%, respectively.
These results follow from our aging formulation, which incorporates temperature using the 
scaling parameter \ineq{\alpha} in (\ref{eq:eq3}) for TDDB and the parameter \ineq{g_0} in (\ref{eq:eq10_nbti}) for NBTI. These parameters grow exponentially with temperature, resulting in a corresponding exponential increase in the aging. Higher aging leads to lower lifetime. 

\subsection{Diode vs. Transistor-based PCM}\label{sec:res_diode_bjt}
\begin{figure}[h!]
	\centering
	\vspace{-5pt}
	\centerline{\includegraphics[width=0.99\columnwidth]{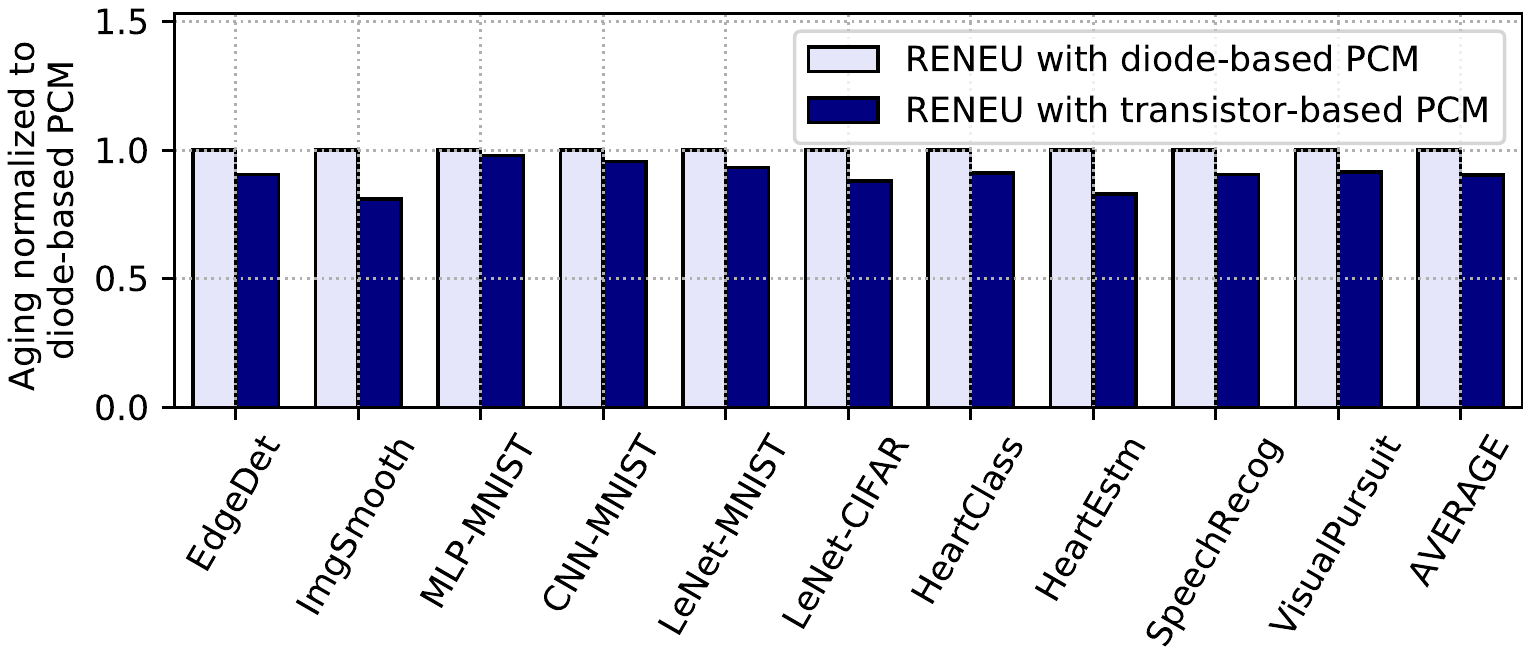}}
	\caption{Circuit aging of \tech{} with transistor-based PCM normalized to \tech{} with diode-based PCM.}
	\vspace{-10pt}
	\label{fig:access}
\end{figure}

Figure~\ref{fig:access} 
reports the circuit aging in a neuromorphic hardware with transistor-based PCM normalized to \tech{} which uses diode-based PCM. We report results for each of our machine learning applications.
We observe that the aging of a neuromorphic hardware with transistor-based PCM is on average 10\% lower than diode-based PCM. This is because the operating voltages of a neuromorphic hardware are comparatively lower for transistor-based PCM, which reduces the circuit aging. However, a diode-based PCM cell is 33\% smaller than a transistor-based PCM cell, which means that diode-based PCM cells can implement neuromorphic hardware with high integration density. Nevertheless, our approach \tech{} can be applied to improve reliability of neuromorphic hardware with both diode and transistor-based PCM.


\subsection{Performance Impact}
\begin{figure}[h!]
	\centering
	\vspace{-5pt}
	\centerline{\includegraphics[width=0.99\columnwidth]{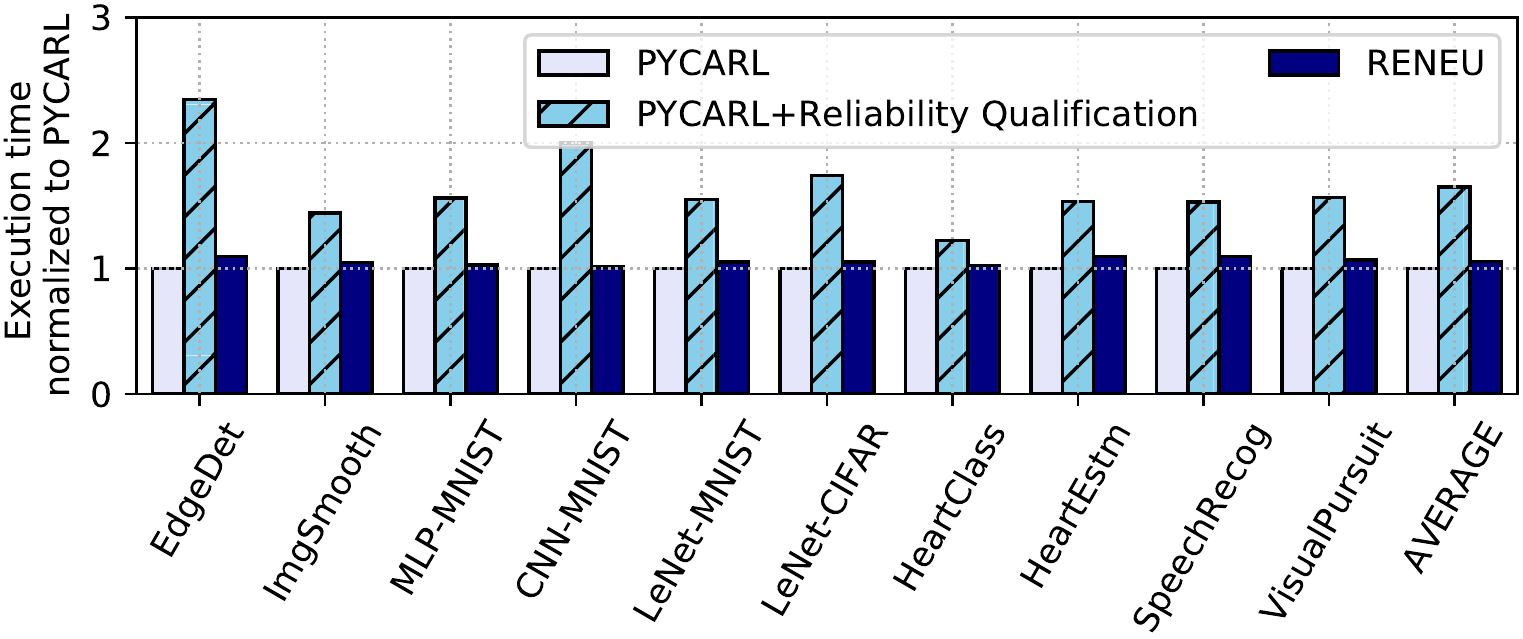}}
	\caption{Execution time of \tech{} normalized to \prior{} (lower is better).}
	\vspace{-10pt}
	\label{fig:extime}
\end{figure}

Figure~\ref{fig:extime} reports the performance of \tech{} measured as the execution time normalized to \prior{} for each of our machine learning applications. We also report the results of a conservative reliability qualification technique, which periodically de-stresses a neuron circuit to achieve similar MTTF as \tech{}~\cite{balaji2019framework}. We make the following two observations. First, the execution time of a machine learning application using \tech{} is within 5\% of the execution time of \prior{}. This is because \tech{} incorporates both performance and reliability when finding a suitable mapping of neurons and synapses to the neuromorphic hardware. 
Second, to achieve a similar MTTF as \tech{}, existing conservative flavor of \prior{} periodically de-stresses the neuron circuit, which introduces an average performance overhead of 35\%.


\subsection{Circuit Aging with More Crossbars}
\begin{figure}[h!]
	\centering
	\vspace{-5pt}
	\centerline{\includegraphics[width=0.99\columnwidth]{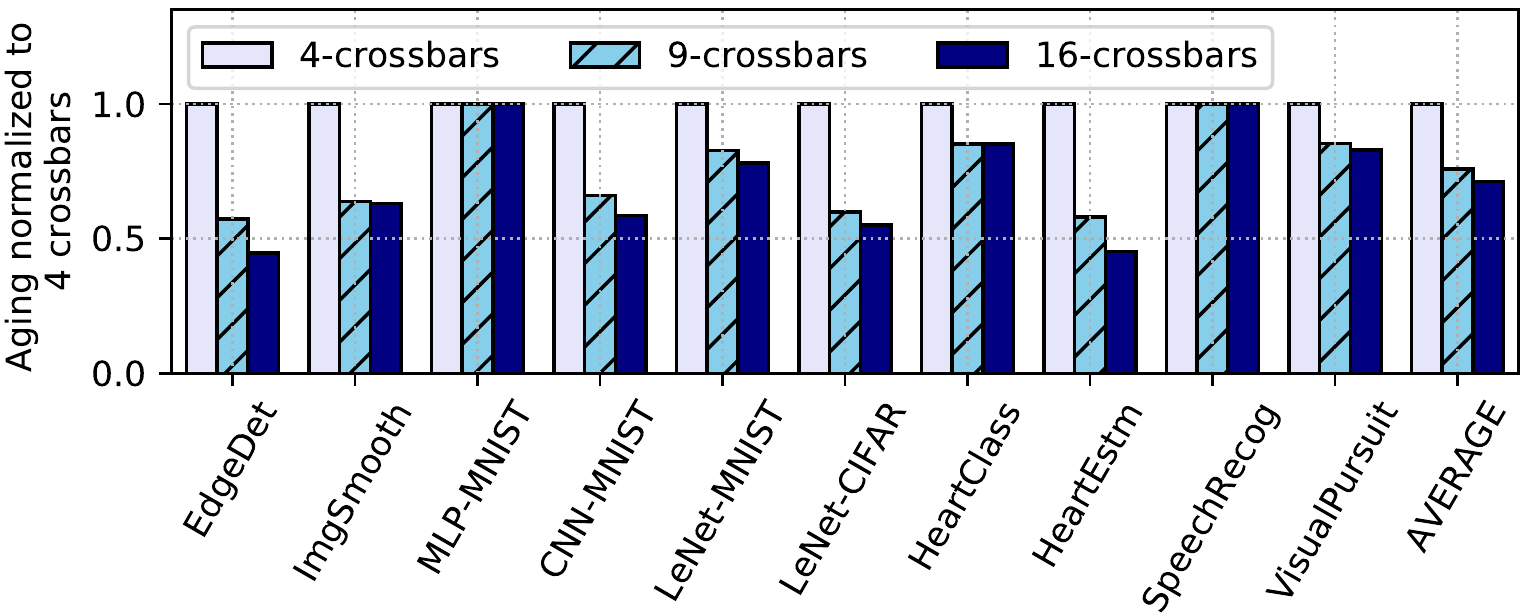}}
	\caption{Circuit aging of \tech{} with 9 and 16 crossbars normalized to \tech{} with 4-crossbars (lower is better).}
	\vspace{-10pt}
	\label{fig:platform}
\end{figure}

Figure~\ref{fig:platform} reports the circuit aging as the amount of hardware resources are increased. We report the aging results of \tech{} with 9 and 16 crossbars for each of our machine learning applications. Aging results are normalized to \tech{} with 4 crossbars. We make the following two observations. First, aging reduces with increasing number of crossbars. This is because with more crossbars in the system, the average load on each crossbar reduces, which in turn reduces the stress on its CMOS devices. Lower stress reduces the circuit aging. With 9 and 16 crossbars, the average circuit aging is respectively, 24\% and 29\% lower than \tech{} with 4 crossbars. Second, for MLP-MNIST, there is no noticeable improvement. This is because MLP-MNIST is a small application to begin with. 

\subsection{Optimization Time}
Table~\ref{tab:opt_distribution} reports the optimization time (measured as wall clock time) of \tech{} in finding a mapping using the proposed PSO. The optimization time depends on the size of the application. The MLP-MNIST, which is a small application requires 4.6s, while LeNet-CIFAR requires 98.2s.

\begin{table}[h]
\renewcommand{\arraystretch}{1.5}
\setlength{\tabcolsep}{2pt}
\caption{Optimization time of \tech{}.}
\label{tab:opt_distribution}
\centering
{\fontsize{9}{10}\selectfont
\begin{tabular}{|l|c|l|c|}
\hline
\textbf{Application} & \textbf{Time (s)} & \textbf{Application} & \textbf{Time (s)}\\
\hline
EdgeDet & 53.1 & ImgSmooth & 47.2 \\
MLP-MNIST & 4.6 & CNN-MNIST & 84.4 \\
LeNet-MNIST & 70.12 & LeNet-CIFAR & 98.2 \\
HeartClass & 14.6 & HeartEstm & 59.93\\
SpeechRecog & 82.0 & Visual Pursuit & 90.8\\
\hline
\end{tabular}}
\end{table}

\vspace{-5pt}



%% file: sections/conclusion.tex
We present \tech{}, a reliability-oriented approach for mapping neurons and synapses to the hardware resources of a Non-volatile Memory (NVM)-based neuromorphic hardware. Prior efforts in mapping neurons and synapses have mostly considered performance. \tech{} is built on two key contributions. \tech{} formulates the detailed circuit aging in a neuromorphic hardware considering different failure mechanisms such as Time Dependent Dielectric Breakdown (TDDB), Negative Bias Temperature Instability (NBTI), and Hot Carrier Injection (HCI). Using this formulation \tech{} places the neurons and synapses to the hardware using an instance of Particle Swarm Optimization (PSO), exploring the performance and reliability trade-offs.
We evaluate \tech{} using machine learning applications on a state-of-the-art neuromorphic  hardware with PCM synapses. Results demonstrate a significant improvement in reliability of neuromorphic computing with marginal impact on performance.